\PassOptionsToPackage{numbers,sort&compress}{natbib}
\documentclass{article}
\usepackage[preprint]{neurips_2026}

\usepackage[utf8]{inputenc}
\usepackage[T1]{fontenc}
\usepackage{hyperref}
\usepackage{url}
\usepackage{booktabs}
\usepackage{amsfonts}
\usepackage{amsmath}
\usepackage{amssymb}
\usepackage{nicefrac}
\usepackage{microtype}
\usepackage{xcolor}
\usepackage{colortbl}
\usepackage{graphicx}
\usepackage{multirow}
\usepackage{longtable}
\usepackage{tabularx}
\usepackage{array}
\usepackage{makecell}
\usepackage{enumitem}
\usepackage{float}
\usepackage{subcaption}
\usepackage{xspace}
\usepackage{tikz}
\usetikzlibrary{arrows.meta,backgrounds,calc,fit,positioning,shapes.geometric,shapes.misc}

\newcommand{\method}{\textsc{Neuro-Symbolic Drive}\xspace}
\newcommand{\dataset}{\textsc{NSD-Sim}\xspace}

\newcommand{\reasoningbegin}{\texttt{<REASONING>}}
\newcommand{\reasoningend}{\texttt{</REASONING>}}
\newcommand{\planningbegin}{\texttt{<PLANNING>}}
\newcommand{\planningend}{\texttt{</PLANNING>}}
\newcommand{\coloneqq}{\mathrel{:=}}

\newcolumntype{Y}{>{\centering\arraybackslash}X}
\newcolumntype{L}[1]{>{\raggedright\arraybackslash}p{#1}}
\newcolumntype{C}[1]{>{\centering\arraybackslash}p{#1}}

\NewDocumentCommand{\tbd}{o}{%
  \begingroup
  \setlength{\fboxsep}{1pt}%
  \colorbox{gray!20}{\textcolor{gray!70!black}{\scriptsize TBD}}%
  \IfValueT{#1}{\textcolor{gray!70!black}{\scriptsize~(#1)}}%
  \endgroup
}

\title{Neuro-Symbolic Drive: Rule-Grounded Faithful Reasoning for Driving VLAs}

\author{%
  Xiangbo Gao$^{1*}$ \quad
  Xiukun Huang$^{2*\dagger}$ \quad
  Boyu Lu$^{3}$ \quad
  Junge Zhang$^{4}$ \\[4pt]
  \bf Mengjie Mao$^{5}$ \quad
  Jiachen Li$^{4}$ \quad
  Wei Xiong \quad
  Zhengzhong Tu$^{1}$ \\[6pt]
  $^1$Texas A\&M University \quad
  $^2$Carnegie Mellon University \quad
  $^3$University of Maryland \\[2pt]
  $^4$University of California, Riverside \quad
  $^5$University of Pittsburgh \\[6pt]
  {\small $^*$Equal contribution \quad $^\dagger$Project lead}
}

\begin{document}

\maketitle

\begin{abstract}
Driving VLA models incorporating Chain-of-Thought (CoT) reasoning are attractive because they leverage pretrained VLM representations and expose intermediate decisions in natural language, yet current rationales often lack the step-by-step decision semantics needed to keep the rationale causally connected to the planned motion. We introduce \method, a neuro-symbolic driving framework that supervises a driving VLA with rule-grounded reasoning traces extracted directly from classical rule-based planners. Our key observation is that rule-based planners are symbolic AI systems that already function as executable reasoning engines: they reason about active safety constraints, search over candidate maneuvers, and select a final trajectory. We instrument these planners in simulation to capture both the executed trajectory and the internal decision trace at each rule-evaluation step. Each trace is serialized into structured rule-grounded reasoning and paired with the trajectory to fine-tune Qwen3.5-4B as a driving VLA. Because these traces are derived directly from the planner states that determine the action, they ensure reasoning is structurally coupled to motion generation by construction, rather than by post-hoc alignment. On our simulator-generated benchmark, detailed rule-grounded reasoning reduces ADE@3s from 0.47 to 0.26 and miss rate from 8.30\% to 6.40\% under three-camera perception, and from 0.54 to 0.26 and 10.13\% to 5.99\% under eight-camera perception. \method thus converts neuro-symbolic planning logic into structured supervision. All source code is publicly available.\footnote{Reasoning trace generation: \url{https://github.com/zcxgi/nuplan-reason}; Model training: \url{https://github.com/XiangboGaoBarry/Neural-Symbolic-Drive}.}
\end{abstract}

\section{Introduction}
\label{sec:introduction}
Classical rule-based planners and end-to-end models represent two complementary paradigms in autonomous driving: the former, which depends on deep perception modules to interpret raw sensor data, offers inspectable decision procedures and explicit safety constraints, but is brittle and limited by perception capabilities; the latter learns flexible representations directly from data but operates as a black box, lacking interpretability and explicit safety constraints.
% Autonomous driving requires an embodied agent that can observe a dynamic scene, reason about route context, traffic rules, and interactions, and commit to a safe future motion.
Before the rise of VLAs, interpretability was largely engineered into the planning stack through structured intermediate representations fed by deep perception, cost volumes, semantic maps, candidate trajectories, and other planning-oriented abstractions \cite{zeng2019nmp, casas2021mp3, uniad, vad, wang2021ivmp, biswas2024quad}. VLA systems offer a different path: by exposing reasoning in natural language, they enable driving agents to articulate scene understanding, interaction logic, and planned behaviors in a form that humans can inspect and audit \cite{chen2023drivingwithllms, sha2023languagempc, sima2024drivelm, xu2023drivegpt4, shao2024lmdrive, hwang2024emma, xing2025openemma, jiang2024senna, tian2025drivevlm, wang2024drivecot, nie2024reason2drive, yang2023llm4drive, cui2024llm4ad, ma2024lampilot, huang2024drivlme, xing2024autotrust, wang2025generative}.

The central difficulty is that readable reasoning is not necessarily action-bearing. Many driving VLAs generate language explanations while predicting trajectories either through a separate regression head or by discretizing continuous actions into tokens appended to the vocabulary. Such designs can produce accurate motions, but offer no guarantee that the stated rationale controls the final trajectory. CoT studies show that plausible explanations can be unfaithful \cite{turpin2023language, lanham2023measuring}, and recent driving work similarly questions whether language reasoning is closely coupled to planning \cite{renz2025simlingo, fu2025orion, li2025driverone, song2025morethanmeetstheeye, tang2026causalvad}. The root cause is a supervision mismatch: current reasoning labels are typically generated post-hoc by humans, other VLMs, or high-level causal annotation pipelines, rather than derived from the mechanism that actually determines the motion. DeepSeek-R1 suggests a principled fix: strong reasoning emerges when intermediate steps are grounded in verifiable, rule-based feedback rather than free-form imitation alone \cite{deepseek2025r1}. We carry this principle into driving: the verifiable feedback is the execution trace of the rule-based planner that determined the trajectory.

Rule-based planners are a natural fit for this role: their internal execution unfolds as a structured decision procedure, enforcing safety rules to prune the action space, scoring competing candidate trajectories, and committing to the best feasible motion. Crucially, these traces are causally responsible for the selected motion, not post-hoc reconstructions. Deployed L4 systems, which already operate driverless fleets across multiple cities worldwide, run precisely such planner stacks: multiple rule-based modules specialized per scenario family, whose decisions are explainable, inspectable, and by construction the causal source of every trajectory they select. Close in spirit, Alpamayo-R1 takes an important step toward decision-grounded supervision by constructing structured Chain-of-Causation (CoC) labels, integrating action generation, and applying post-training alignment to bridge reasoning and trajectory generation for long-tail driving \cite{wang2025alpamayo}. We are inspired by its focus on reasoning-action alignment, but target a different supervision source: rather than relying on human- or VLM-generated causal labels, we extract reasoning directly from the executable rule-based planner.

\begin{figure}[ht]
    \centering
    \includegraphics[width=\textwidth]{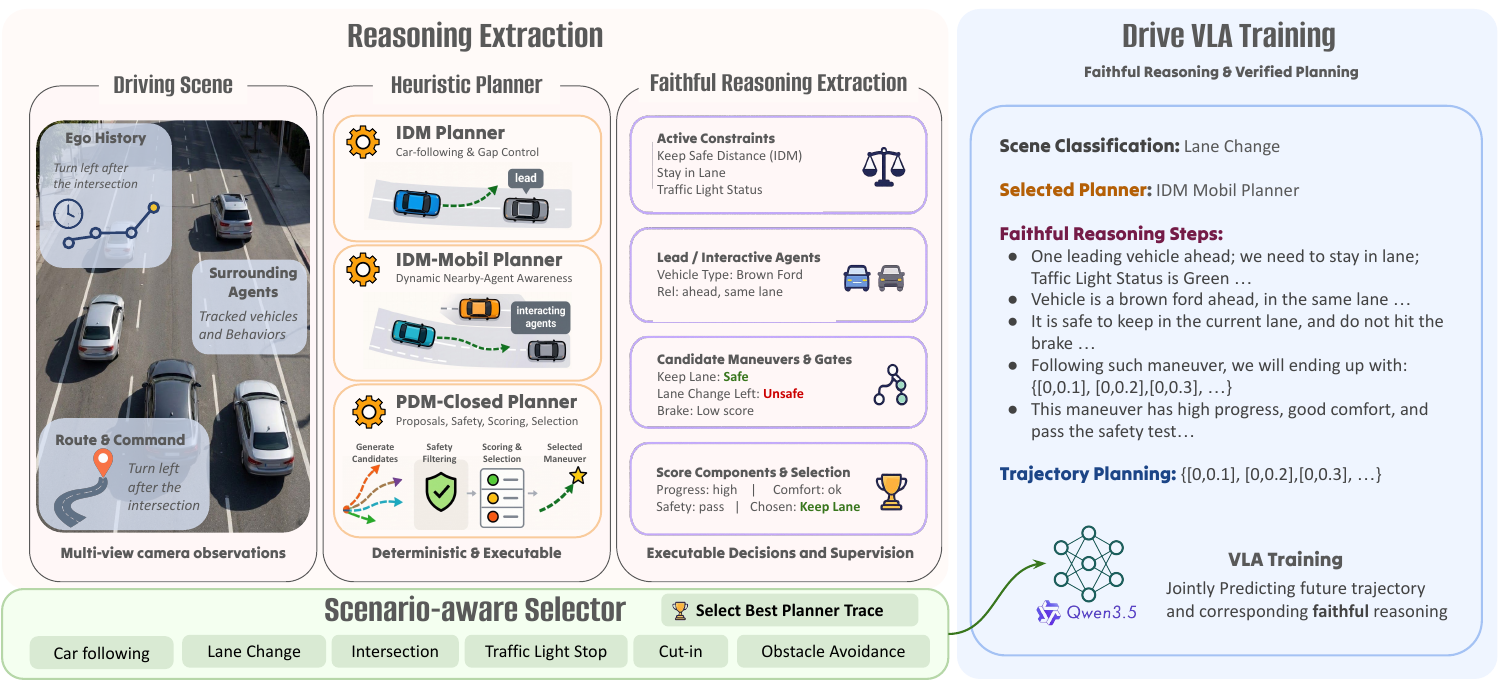}
    \caption{\textbf{Overview of \method.} The left side extracts structured decision traces from rule-based planners. For each simulated driving scene, planner traces expose active constraints, relevant agents, candidate maneuvers, safety gates, score components, and final selections. A scenario-aware selector chooses the trace best matched to the scene family. The right side uses these traces to supervise a driving VLA, so the model learns to jointly produce faithful reasoning and the corresponding future trajectory.}
    \label{fig:nsdrive_overview}
    \vspace{-0.1cm}
\end{figure}

We propose \method, a neuro-symbolic framework that converts rule-based planner execution traces into structured supervision for a driving VLA. Given a simulated scene, we run complementary planners, record their decision traces, select the planner best matched to the scenario family, and convert the chosen trace into rule-grounded reasoning paired with the trajectory. Because the reasoning trace and the trajectory arise from the same planner execution, the rationale is action-bearing by construction, rather than post-hoc alignment. Figure~\ref{fig:nsdrive_overview} summarizes this pipeline.

Our main contributions are as follows.

\begin{itemize}[leftmargin=*, itemsep=2pt, topsep=2pt]
    \item We introduce \method, a neuro-symbolic framework that repurposes rule-based planner execution traces as structured reasoning supervision for vision-language driving agents.
    \item We design a lightweight abstraction that maps heterogeneous raw planner states from multiple teachers to a unified four-slot reasoning schema, enabling cross-teacher learning without planner-specific supervision vocabularies.
    % without adding planner-specific modules to the VLM.
    \item We demonstrate that rule-grounded reasoning improves driving performance and produces more behaviorally consistent rationales than trajectory-only supervision.
\end{itemize}

\section{Related Work}
\label{sec:related}

\subsection{Language Reasoning in Driving VLAs}

% Language has been used in autonomous driving both as an interface to a structured driving state and as part of end-to-end decision making.
Early language-augmented systems translate object vectors, route context, or traffic descriptions into prompts or planning states \cite{chen2023drivingwithllms, sha2023languagempc, wang2023drivemlm}, while recent VLA models directly connect multi-view observations to explanations, decisions, controls, or trajectories \cite{sima2024drivelm, xu2023drivegpt4, shao2024lmdrive, hwang2024emma, xing2025openemma, jiang2024senna, tian2025drivevlm, nie2024reason2drive, wang2024drivecot, yuan2024ragdriver}.
A parallel agentic line studies LLM and VLM driving agents that must follow instructions, reason about rules, use experience, and act under interaction \cite{yang2023llm4drive, cui2024llm4ad, gao2025automated, ma2024lampilot, huang2024drivlme, gao2025langcoop, godbole2025drama}, and recent VLA systems further explore adaptive reasoning, RL, instructed action generation, and decoupled high-level reasoning \cite{zhou2025autovla, li2025driverone, ye2026autodrivep3, fu2025orion, tao2026navidrivevlm}.
These works show language can make driving agents more interpretable, but many still rely on free-form rationales or post-hoc alignment; \method instead derives reasoning directly from the executable planner state that produced the trajectory.

\subsection{Chain-of-Thought Reasoning, Faithfulness, and the Supervision Mismatch}

Inference-time scaling, from CoT \cite{wei2022cot} to process reward models \cite{lightman2024lets}, consistently shows that step-level supervision can be more effective than outcome supervision, with OpenAI o1 \cite{OpenAI2024o1} and DeepSeek-R1 \cite{deepseek2025r1} identifying rule-based verifiable rewards as a key driver over free-form imitation.
% Inspired, the autonomous driving community has adopted CoT in VLA models.
EMMA \cite{hwang2024emma} represents continuous control signals as discrete tokens using the existing VLM's pretrained tokenizer, so driving actions and language reasoning share the same word embeddings and semantic space.
DriveVLM \cite{tian2025drivevlm} uses language in a similar way as an intermediate reasoning layer between perception and trajectory generation; DriveLM \cite{sima2024drivelm} structures reasoning as a graph of visual question-answering steps; DriveCoT \cite{wang2024drivecot} and Reason2Drive \cite{nie2024reason2drive} generate supervised thinking-process traces from expert demonstrations.
Alpamayo-R1 \cite{wang2025alpamayo} constructs structured CoC labels and applies post-training alignment to bridge decision-grounded reasoning and trajectory generation.
% This critique connects to a broader faithfulness problem.
CoT studies warn that plausible explanations can omit or distort the true factors behind a model output \cite{turpin2023language, lanham2023measuring}, and imitation policies may exploit correlated but non-causal signals \cite{dehaan2019causal}.
In autonomous driving, these concerns surface concretely as language-action mismatch, weak reasoning-planning coupling, and confounding in end-to-end planning \cite{renz2025simlingo, fu2025orion, li2025driverone, song2025morethanmeetstheeye, tang2026causalvad}.
The root cause is a supervision mismatch: existing approaches generate reasoning labels post-hoc, as text narrations, causal annotations, or latent surrogates, rather than from the mechanism that determined the motion.
\method addresses this at the source: reasoning traces are abstracted directly from the planner computation that selected the supervision trajectory, not constructed after the fact.

\subsection{Rule-Based Planners and Decision-Grounded Supervision}

Interpretable planning precedes current driving VLAs.
Neural planners and map-perceive-predict-plan systems expose detections, semantic maps, and cost volumes as structured intermediate variables \cite{zeng2019nmp, casas2021mp3}; end-to-end planning-oriented systems retain task decomposition \cite{uniad, vad}; and IVMP and QuAD emphasize interpretable factors and candidate evaluation \cite{wang2021ivmp, biswas2024quad}.
Closed-loop benchmarks show that rule-based planners remain competitive and complementary to learned models \cite{caesar2021nuplan, dauner2023misconceptions, ghosh2026radlad}.
What this body of work does not exploit is that rule-based planners produce a rich internal execution state comprising active constraints, candidate proposals, safety checks, and score components, all causally synchronized with the selected trajectory.
On the supervision side, decision-grounded learning has appeared in several forms: DriveMLM aligns language output with planning states \cite{wang2023drivemlm}, DriveCoT generates thinking-process labels from rule-based experts \cite{wang2024drivecot}, Hydra-MDP distills from human and rule-based teachers \cite{li2024hydramdp}, and DiMA distills LLM knowledge into efficient driving models \cite{hegde2025dima}.
Program-grounded learning further suggests that executable structure can be serialized and distilled into neural models \cite{shlapentokhrothman2025vpd}, while NaviDriveVLM and RAD-LAD explore decoupled or hybrid language-planner systems \cite{tao2026navidrivevlm, ghosh2026radlad}.
Across all of these lines, planners are treated as baselines, safety modules, or inference-time collaborators, and reasoning labels are generated or aligned after the trajectory is determined.
\method differs by treating planner execution traces as the primary supervision source: the reasoning and the trajectory arise from the same transparent computation, making the rationale action-bearing by construction rather than by post-hoc alignment.

\section{Method}
\label{sec:method}

\begin{figure}[t]
    \centering
    \includegraphics[width=\textwidth]{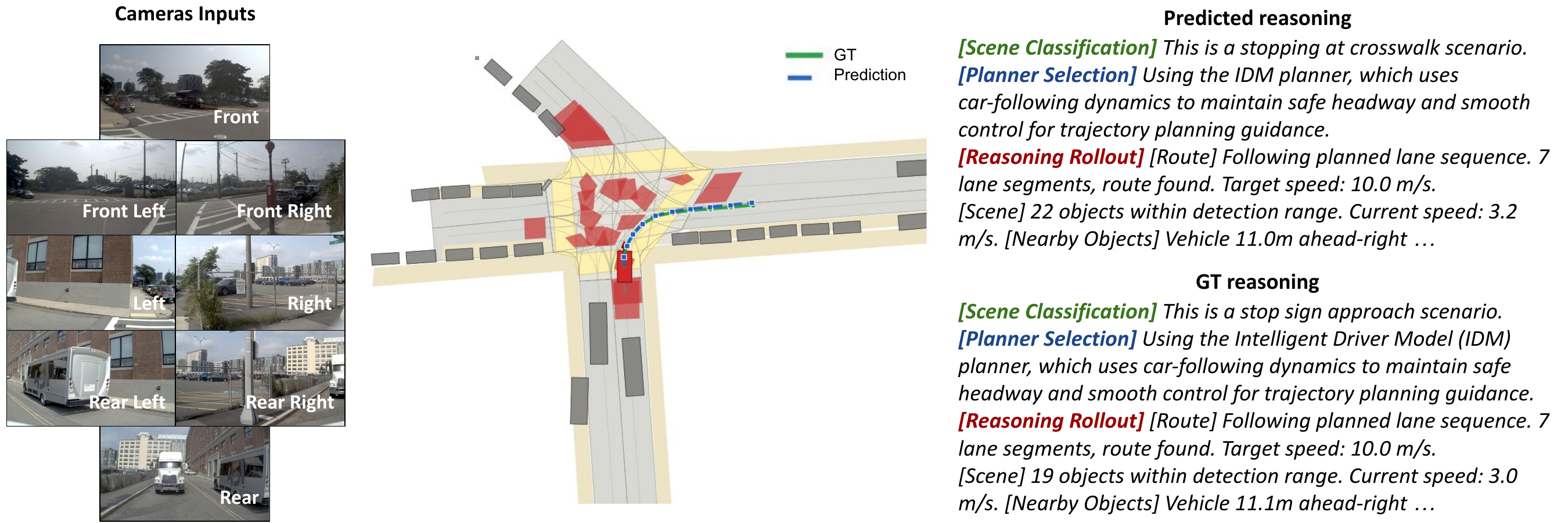}
    \caption{\textbf{Generated reasoning and trajectory visualization.} We show a qualitative prediction from the trained model with the generated rule-grounded reasoning trace and the corresponding top-down trajectory comparison. The example illustrates that the model does not merely output waypoints: it first identifies the active driving context and decision rationale, then predicts a trajectory that is visually aligned with the ground-truth motion.}
    \label{fig:qual_reasoning}
    \vspace{-0.1cm}
\end{figure}

We treat the faithfulness gap, the disconnect between a model's stated rationale and what actually determined its motion, as a supervision sourcing problem: reasoning labels should be derived from the mechanism that determined the motion, not constructed after the fact.

\subsection{Problem Formulation}
\label{sec:problem}

At timestep $t$, the driving input is
\begin{equation}
x_t \coloneqq \left(I_t, h_t, g_t\right),
\end{equation}
where $I_t$ denotes synchronized multi-view images, $h_t$ is the recent ego-state history, and $g_t$ is the route or mission context. A deterministic planner $p \in \mathcal{P}$ (the set of available rule-based planners) receives $x_t$ and returns both a future trajectory $y_t$ and an internal execution trace $z_t$:
\begin{equation}
(y_t, z_t) = p(x_t).
\label{eq:planner_pair}
\end{equation}

The trace $z_t$ is the raw program state comprising active constraints, route context, candidate proposals, safety checks, score components, and final trajectory choice, not free-form text. An abstraction operator $\mathcal{A}$ converts it into a reasoning sequence $r_t$:
\begin{equation}
r_t \coloneqq \mathcal{A}(z_t).
\label{eq:abstraction}
\end{equation}

The training target for the VLA is the concatenated sequence
\begin{equation}
u_t \coloneqq [\reasoningbegin \; r_t \; \reasoningend \; \planningbegin \; y_t \; \planningend].
\label{eq:joint_target}
\end{equation}
The key design choice is that $r_t$ is derived from the same planner state that produced $y_t$, making the reasoning action-bearing by construction rather than merely correlated with the trajectory.

\subsection{Planner Trace Extraction}
\label{sec:trace}

We use three complementary rule-based planners in simulation: Intelligent Driver Model Planner (IDM), IDM-MOBIL Planner, which augments IDM with the MOBIL lane-change criterion (Minimizing Overall Braking Induced by Lane Changes)~\cite{10.1007/978-3-540-77074-9_19}, and Predictive Driver Model Closed Planner (PDM-Closed)~\cite{dauner2023misconceptions}. IDM provides stable centerline car-following traces through lead-agent, headway, closing-speed, safety-gap, and acceleration terms, while IDM-MOBIL augments IDM with interaction-aware lane-change and gap-selection logic for cut-in, merge, queueing, and close-following cases. PDM-Closed follows a proposal-based closed-loop procedure: it generates candidate trajectories, applies safety and progress checks, scores surviving proposals, and selects the final maneuver~\cite{10803052, dauner2023misconceptions, karnchanachari2024nuplan}. These three teachers therefore expose complementary reasoning patterns: gap control, interaction-aware maneuver selection, and propose-evaluate-select planning.

\begin{figure}[t]
    \centering
    \includegraphics[width=\textwidth]{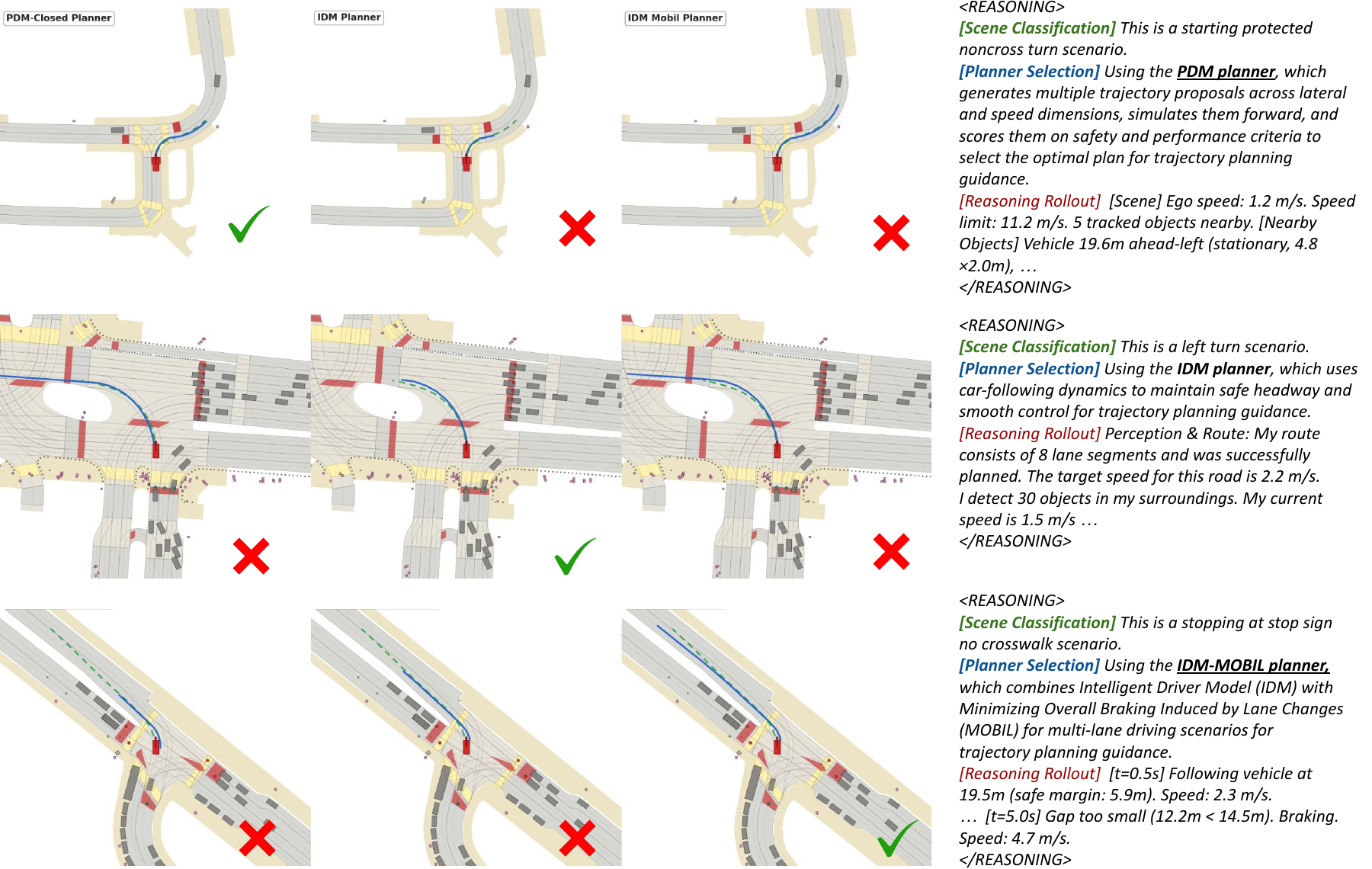}
    \caption{\textbf{Same-scene planner comparison.} We visualize the three rule-based teachers on matched scenarios and timestamps. The examples show that different rule-based planners can make distinct closed-loop choices under the same observation, which motivates selecting the teacher trace according to scenario-level closed-loop quality rather than treating any single planner as a universal expert.}
    \label{fig:planner_compare}
    \vspace{-0.1cm}
\end{figure}

As Figure~\ref{fig:planner_compare} shows, rather than relying on a single universal teacher, we route each scenario family to its most capable planner: closed-loop evaluation identifies the most reliable teacher per family, whose trace and trajectory become the default supervision. This teacher-selection step is the data-construction analog of L4 dispatch. When multiple planners are complementary, we also retain multiple successful rollouts for the same scene, increasing supervision diversity without forcing a single teacher to cover scenarios outside its competence.

\begin{figure}[t]
    \centering
    \vspace{-3mm}
    \includegraphics[width=\textwidth]{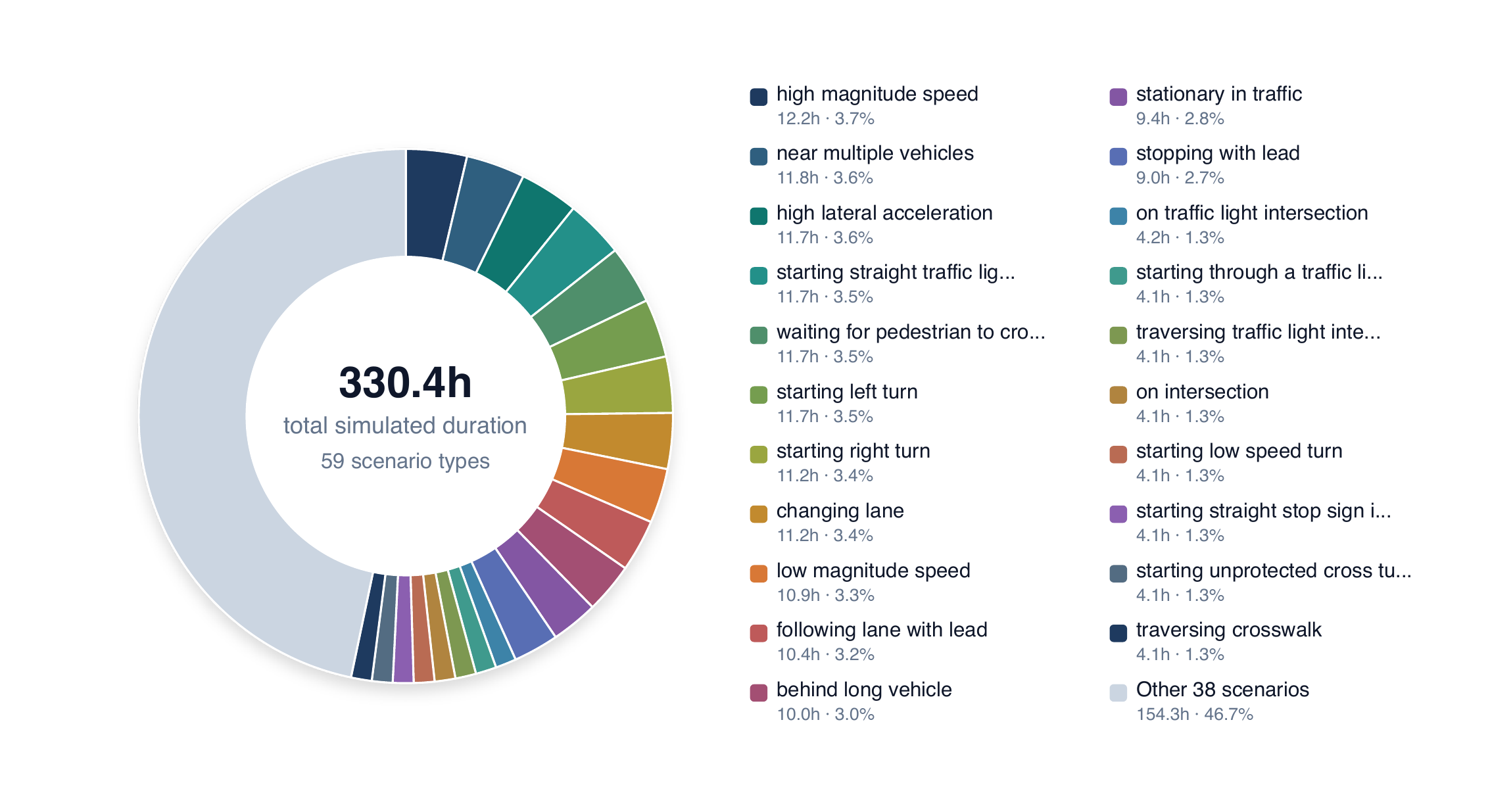}
    \vspace{-13mm}
    \caption{\textbf{Merged dataset duration by scenario type.} We compute each scenario's simulated duration as the number of examples multiplied by the average scenario duration. The chart shows the top 21 scenario types and groups the remaining 38 types as \emph{Other}, illustrating the diverse scenario coverage used to extract rule-grounded reasoning traces.}
    \label{fig:dataset_duration}
    \vspace{-3mm}
\end{figure}

Figure~\ref{fig:dataset_duration} summarizes the merged scenario coverage used for planner trace extraction. The distribution spans 59 scenario types and roughly 330 hours of simulated duration in the dataset, with the largest individual scenario families contributing only a small fraction of the full corpus. This broad but relatively balanced coverage is important for our setting: the model sees planner traces from routine lane following, longitudinal car-following, traffic-light traversal, turning, interaction, and long-tail maneuver scenarios, rather than learning reasoning from a narrow planner-specific subset.

Instrumentation is read-only: we record the planner program state without changing the trajectory computation itself. The extracted fields include route context, relevant agents, safety constraints, candidate outcomes, score components, selected proposal identity, and fallback status, when applicable. Because the trace is captured from the same execution that produces the trajectory, every reasoning target remains synchronized with the exact motion used for supervision.

\subsection{Reasoning Abstraction and Serialization}
\label{sec:serialization}

Raw program traces are not suitable supervision targets by themselves. First, the three planners expose different internal vocabularies: IDM is organized around lead vehicles, headway, desired gaps, and acceleration updates; IDM-MOBIL adds lateral incentives and lane-change safety checks; PDM-Closed exposes proposal banks, safety filters, and score components. A single VLA cannot learn unified decision semantics from these heterogeneous variable names alone. Second, raw traces mix action-relevant evidence with implementation artifacts such as loop counters, temporary buffers, and intermediate accumulators. Directly imitating them would encourage memorization of planner-specific execution details rather than learning reusable driving logic. Figure~\ref{fig:qual_reasoning} provides a concrete example of the resulting serialized reasoning trace together with the corresponding top-down generated trajectory.
We therefore serialize each trace into a compact schema with four slots:
\begin{equation}
r_t \coloneqq \left[s_t^{\text{scene}}, s_t^{\text{constraint}}, s_t^{\text{candidate}}, s_t^{\text{decision}}\right].
\end{equation}
The \emph{scene} summarizes the route and ego context. The \emph{constraint} specifies the active safety or rule signal. The \emph{candidate} describes either the gap update sequence or the proposal-screening outcomes. The \emph{decision} states the chosen maneuver and its immediate consequence for future motion.

We instantiate two text realizations. The \emph{concise} variant keeps the schema short and quantitative, exposing \emph{what} the planner decided at each step and yielding a stable supervision target with limited token overhead. The \emph{detailed} variant expands the same slots into a step-by-step narrative that exposes \emph{how} each decision was reached through candidate filtering and rejection rationale. Both variants share the same four-slot semantics; they differ only in the depth of the decision trace exposed. This lets us test whether performance gains come from the decision structure or the depth of the explanation.

The target sequence is serialized as rule-grounded reasoning first, followed by the future trajectory (Eq.~\ref{eq:joint_target}). Explicit delimiters are used in the implementation, but serialized targets exclude planner-specific method names and simulator-specific debug fields. The reasoning sequence only preserves decision semantics that remain stable between teachers and scenarios.

\section{Experiments}
\label{sec:experiments}

The goal of our experiments is to test whether rule-grounded reasoning improves autonomous-driving planning, rather than to rank rule-based planners as standalone systems. Because our supervision is extracted from simulator rollouts, all training, inference, and testing are conducted in simulation. This gives us synchronized observations, planner traces, and target trajectories, and therefore a controlled setting for isolating the effect of reasoning depth and teacher selection.

\subsection{Experimental Setup}
\label{sec:exp_setup}

We evaluate on the \dataset dataset, a simulator-generated corpus built from 520 nuPlan scenarios and 77{,}558 timestep-level training examples. Each example contains synchronized multi-view images, ego history, mission context, a planner-generated trajectory, and the corresponding planner trace. We use a scenario-level split to avoid timestep leakage, with 70\% of scenarios for training, 15\% for validation, and 15\% for testing. We evaluate two camera settings: a 3-camera setting with front, front-left, and front-right views, and an 8-camera setting with the full surrounding camera suite. Following common planning evaluation protocols in autonomous driving \cite{caesar2020nuscenes, caesar2021nuplan, uniad, vad, shao2024lmdrive, hwang2024emma, xing2025openemma}, we evaluate the generated future trajectory against the planner-generated target trajectory (which serves as ground truth in our simulation setting) using Average ADE, ADE/FDE at 1s/2s/3s, AHE/FHE at 3s, and Miss Rate at 3s. ADE/FDE measure displacement error, AHE (Average Heading Error) / FHE (Final Heading Error) measure heading error, and Miss Rate measures the fraction of generations that exceed the predefined trajectory error threshold.

We instantiate the driving backbone with Qwen3.5-4B \cite{qwen35blog} and fine-tune all parameters with standard supervised learning. Unless otherwise noted, all variants use the same optimizer, tokenizer, image preprocessing, maximum sequence length, and number of epochs. The controlled variables are the camera setting, the reasoning level, and the planner source used to construct the training data. We compare the three supervision levels defined in Section~\ref{sec:serialization} — without reasoning, concise, and detailed — to test whether richer planner decision exposure improves trajectory generation, and whether gains come from decision structure or explanation depth. Since all conditions share the same trajectory targets, any improvement reflects the supervisory signal rather than the reference motion.

\subsection{Main Results: Camera Setting and Reasoning Depth}
\label{sec:exp_main}

\begin{table*}[ht]
\centering
\caption{\textbf{Main planning results.} All rows use the full multi-planner data construction pipeline. The comparison isolates whether rule-grounded reasoning depth improves planning performance under 3-camera and 8-camera perception settings. Lower is better for all metrics.}
\label{tab:main_results}
\small
\setlength{\tabcolsep}{3.2pt}
\renewcommand{\arraystretch}{1.12}
\resizebox{\textwidth}{!}{%
\begin{tabular}{lcccccccccc}
\toprule
Reasoning level
& Average ADE$\downarrow$
& ADE@1s$\downarrow$ & ADE@2s$\downarrow$ & ADE@3s$\downarrow$
& FDE@1s$\downarrow$ & FDE@2s$\downarrow$ & FDE@3s$\downarrow$
& AHE@3s$\downarrow$ & FHE@3s$\downarrow$
& MR@3s$\downarrow$ \\
\midrule
\multicolumn{11}{c}{\textit{Three Cameras}} \\
\midrule
Without reasoning & 0.32 & 0.22 & 0.33 & 0.47 & 0.21 & 0.47 & 0.68 & 1.99 & 1.85 & 8.30\% \\
Concise reasoning & 0.20 & 0.13 & 0.19 & 0.28 & \textbf{0.14} & 0.29 & 0.52 & 1.04 & 1.05 & 6.90\% \\
\rowcolor{blue!6}
Detailed reasoning & \textbf{0.19} & \textbf{0.12} & \textbf{0.18} & \textbf{0.26} & \textbf{0.14} & \textbf{0.27} & \textbf{0.49} & \textbf{1.02} & \textbf{1.01} & \textbf{6.40\%} \\
\midrule
\multicolumn{11}{c}{\textit{Eight Cameras}} \\
\midrule
Without reasoning & 0.36 & 0.24 & 0.31 & 0.54 & 0.28 & 0.44 & 0.88 & 1.59 & 1.48 & 10.13\% \\
Concise reasoning & 0.20 & \textbf{0.13} & 0.18 & 0.27 & \textbf{0.13} & 0.28 & 0.49 & 1.03 & 0.97 & 6.50\% \\
\rowcolor{blue!6}
Detailed reasoning & \textbf{0.19} & \textbf{0.13} & \textbf{0.17} & \textbf{0.26} & \textbf{0.13} & \textbf{0.27} & \textbf{0.46} & \textbf{0.99} & \textbf{0.92} & \textbf{5.99\%} \\
\bottomrule
\end{tabular}
}
\vspace{-0.1cm}
\end{table*}

\paragraph{Main results.} Table~\ref{tab:main_results} shows a clear and consistent trend: adding rule-grounded reasoning improves planning quality under both perception settings, and richer reasoning consistently yields the best or tied results. In the three-camera setting, concise reasoning already produces a large gain over the no-reasoning baseline, reducing Average ADE from 0.32 to 0.20, ADE@3s from 0.47 to 0.28, and MR@3s from 8.30\% to 6.90\%. Detailed reasoning improves the same model further, lowering Average ADE to 0.19, ADE@3s to 0.26, FDE@3s to 0.49, and MR@3s to 6.40\%. This pattern suggests that rule-grounded reasoning helps not only on long-horizon failure cases, but also on the overall trajectory quality summarized by Average ADE.

The eight-camera setting reinforces the same pattern. The no-reasoning baseline degrades to 0.36 Average ADE and 10.13\% MR@3s, showing that more views do not by themselves produce stable planning. Once rule-grounded reasoning is added, performance improves sharply: concise reasoning reduces Average ADE to 0.20 and MR@3s to 6.50\%, while detailed reasoning reaches the best overall result with 0.19 Average ADE, 0.26 ADE@3s, 0.46 FDE@3s, 0.99 AHE@3s, 0.92 FHE@3s, and 5.99\% MR@3s. The gain does not come from a richer sensor suite; it comes from intermediate decision semantics that organize how perceptual evidence translates into behavior.

Comparing concise and detailed reasoning is also informative. Concise reasoning, which exposes \emph{what} the planner decided at each step, already captures most of the improvement, indicating that active constraints and selected maneuvers carry the highest-value supervisory signal. Detailed reasoning, which additionally exposes \emph{how} each decision was reached through candidate filtering and rejection rationale, still brings a further, repeatable gain on long-horizon displacement, heading, and miss-rate metrics, implying that step-by-step decision semantics provide additional information beyond the final maneuver label. Taken together, these results support the central thesis: reasoning helps not because it makes the model more verbose, but because rule-grounded reasoning makes the planning target easier to align with and learn from.

Figure~\ref{fig:qual_reasoning} gives a qualitative view of the same trend. The generated reasoning trace exposes the scene-level constraint and maneuver choice before the model emits future waypoints, and the top-down overlay makes it possible to inspect whether the planned motion is consistent with that reasoning.

\subsection{Ablation: Planner Sources and Data Organization}
\label{sec:exp_planner_ablation}

\begin{table*}[ht]
\centering
\caption{\textbf{Planner-source ablation.} All rows use detailed rule-grounded reasoning. The comparison varies only which rule-based teacher supplies reasoning traces and trajectories, and how multiple teachers are selected or organized. Lower is better for all metrics.}
\label{tab:planner_ablation}
\small
\setlength{\tabcolsep}{2.6pt}
\renewcommand{\arraystretch}{1.12}
\resizebox{\textwidth}{!}{%
\begin{tabular}{llcccccccccc}
\toprule
Teacher source & Selection policy
& Average ADE$\downarrow$
& ADE@1s$\downarrow$ & ADE@2s$\downarrow$ & ADE@3s$\downarrow$
& FDE@1s$\downarrow$ & FDE@2s$\downarrow$ & FDE@3s$\downarrow$
& AHE@3s$\downarrow$ & FHE@3s$\downarrow$
& MR@3s$\downarrow$ \\
\midrule
\multicolumn{12}{c}{\textit{Three Cameras}} \\
\midrule
IDM & single teacher & 0.31 & 0.21 & 0.25 & 0.41 & 0.27 & 0.49 & 0.86 & 1.70 & 1.72 & 10.21\% \\
PDM-Closed & single teacher & 0.28 & 0.15 & 0.26 & 0.40 & 0.22 & 0.39 & 0.57 & 1.30 & 1.39 & 8.47\% \\
IDM-MOBIL & single teacher & 3.17 & 3.04 & 3.45 & 3.60 & 2.62 & 3.61 & 6.02 & 6.65 & 4.81 & 30.16\% \\
All Planners & random teacher per scene & 0.22 & 0.15 & 0.20 & 0.28 & 0.15 & 0.30 & 0.54 & \textbf{1.02} & 1.07 & \textbf{6.27\%} \\
\rowcolor{blue!6}
All Planners & scenario-aware selection & \textbf{0.19} & \textbf{0.12} & \textbf{0.18} & \textbf{0.26} & \textbf{0.14} & \textbf{0.27} & \textbf{0.49} & \textbf{1.02} & \textbf{1.01} & 6.40\% \\
\midrule
\multicolumn{12}{c}{\textit{Eight Cameras}} \\
\midrule
IDM & single teacher & 0.29 & 0.21 & 0.27 & 0.40 & 0.24 & 0.47 & 0.78 & 1.67 & 1.64 & 10.32\% \\
PDM-Closed & single teacher & 0.24 & 0.16 & 0.22 & 0.34 & 0.18 & 0.35 & 0.59 & 1.26 & 1.19 & 7.39\% \\
IDM-MOBIL & single teacher & 3.13 & 2.62 & 3.08 & 3.70 & 2.76 & 3.84 & 5.35 & 6.12 & 5.25 & 28.91\% \\
All Planners & random teacher per scene & 0.20 & \textbf{0.13} & 0.19 & 0.27 & 0.14 & 0.28 & 0.51 & 1.06 & 0.97 & 6.22\% \\
\rowcolor{blue!6}
All Planners & scenario-aware selection & \textbf{0.19} & \textbf{0.13} & \textbf{0.17} & \textbf{0.26} & \textbf{0.13} & \textbf{0.27} & \textbf{0.46} & \textbf{0.99} & \textbf{0.92} & \textbf{5.99\%} \\
\bottomrule
\end{tabular}
}
\vspace{-0.1cm}
\end{table*}

\paragraph{Planner-source ablation.} The single-teacher rows show that no rule-based planner is a universally reliable source of supervision. PDM-Closed is stronger than IDM on most metrics in both camera settings, while IDM-MOBIL alone performs poorly when used as the only teacher. This result is expected rather than a data bug: IDM-MOBIL is primarily designed for car-following with lane-change incentives and nearby-agent interaction, but it has limited ability to handle route-level turning, intersection traversal, and many traffic-control cases. When forced to supervise the entire benchmark, it produces very large errors on those out-of-scope scenes. Its value appears in a narrower set of lateral maneuver and interaction-heavy scenarios, which is why we use it as a specialized teacher inside the multi-planner mixture rather than as a universal teacher.

Multi-teacher data is consistently better than any single-teacher setting: with three cameras, random teacher per scene reduces MR@3s from 8.47\% for the best single teacher to 6.27\%; with eight cameras, scenario-aware selection reaches the best overall result with 0.19 Average ADE and 5.99\% MR@3s. The comparison between the two multi-planner policies is also informative. In the random-teacher setting, each scene still contributes only one retained training example: we first form a pool of valid rollouts from the three planners, then randomly choose one teacher trace and trajectory for that scene. Thus, the dataset size does not grow relative to the single-teacher and scenario-aware settings, and the gain cannot be attributed to a simple increase in supervision volume. Scenario-aware selection is the cleaner policy for our main claim because it assigns one default teacher to each scenario family according to closed-loop reliability, reducing conflicting targets while preserving the same dataset scale. This supports our design choice of using planner capability estimates to decide which reasoning trace and trajectory should supervise each scenario family.

\subsection{Quality of the Rule-based Planner Teachers}
\label{sec:exp_teacher_quality}

\begin{figure}[ht]
    \centering
    \includegraphics[width=\textwidth]{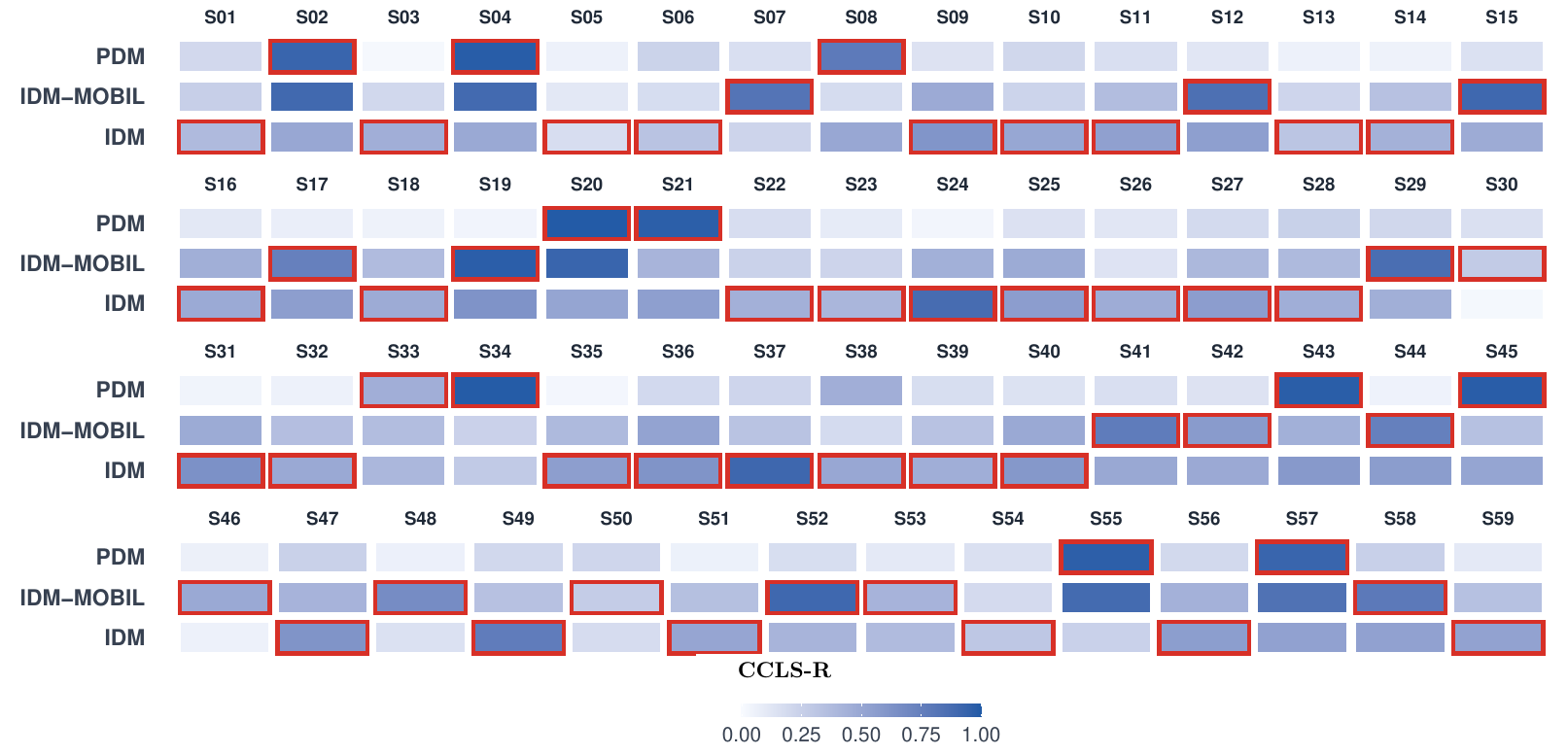}
    \caption{\textbf{Closed-loop teacher quality.} Each indexed column is a shuffled scenario family in the dataset, and each row is one teacher planner. Color indicates CCLS-R (higher is better). Red boxes mark the highest-CCLS-R teacher per scenario, guiding scenario-aware selection of the planner trace and trajectory for VLA supervision. Scenario-index mappings are listed in Appendix~\ref{sec:app_scenario_mapping}.}
    \label{fig:teacher_quality}
    \vspace{-0.1cm}
\end{figure}

Figure~\ref{fig:teacher_quality} reports the closed-loop quality of the rule-based teachers used to construct rule-grounded supervision over the 59 scenario families observed in our dataset. We use the Composite Closed-Loop Score Reactive (CCLS-R) to compare planners across scenario families. For each rollout, CCLS-R combines the major closed-loop safety and progress terms:
\begin{equation}
\begin{aligned}
\mathrm{CCLS-R}\coloneqq{}&
0.35 C_{\mathrm{collision}}
+0.25 C_{\mathrm{drivable}}
+0.15 C_{\mathrm{ttc}} \\
&+0.10 C_{\mathrm{direction}}
+0.10 C_{\mathrm{comfort}}
+0.05 \min(P_{\mathrm{progress}},1).
\end{aligned}
\end{equation}
This metric is intentionally broader than a binary success rate: it captures whether a planner produces usable closed-loop behavior even when nuPlan's strict aggregate score is low. For data construction, CCLS-R ranks teachers within each scenario family. We then use the best-CCLS-R teacher as the default source of reasoning traces and trajectories, after applying the simulator route-success filter. A rollout is retained for supervision only if it completes the full planned trajectory without collision or off-road violation. We do not apply an additional global CCLS-R threshold; CCLS-R guides teacher selection, while the route-success filter determines whether a rollout is kept for VLA training.

Figures~\ref{fig:planner_compare} and~\ref{fig:teacher_quality} support a mixture-of-teachers strategy. PDM-Closed is the strongest generalist in most intersection, turn, and traffic-light scenarios; IDM-MOBIL is more competitive in lateral maneuver and static-obstacle cases; and IDM remains useful in conservative crawl-and-yield settings where partial closed-loop credit matters. Thus, the supervising planner is not chosen arbitrarily: closed-loop evidence determines which teacher is most reliable per scenario family.

\section{Conclusion}
\label{sec:conclusion}

We introduced \method, a neuro-symbolic framework that grounds vision-language driving models in executable planner semantics rather than post-hoc rationalization. The core insight is that rule-based planners already encode rich decision-making logic: constraint activation, proposal filtering, safety validation, and outcome ranking. This internal execution trace can be repurposed as faithful reasoning supervision for training driving VLAs. By instrumenting three complementary planners and routing each scenario to its most capable teacher, mirroring the L4 hierarchy of specialized planners per scenario family, we ensure that reasoning targets stay synchronized with the trajectories they explain. Empirically, detailed rule-grounded reasoning reduces ADE@3s by 45\% and miss rate by 23\% under three-camera perception, and brings an unstable eight-camera baseline to parity with the three-camera setting. Executable decision semantics are a strong supervision signal; more broadly, any expert system with an executable trace, such as a constraint solver or rule engine, is a candidate supervision source.

\paragraph{Limitations.}
Our study operates in simulation: supervision is synthetic, derived from rule-based planner rollouts, and the VLA is evaluated in open-loop rather than deployed in a real-world closed loop. This is intentional: rule-based planner execution traces are not preserved in passively logged sensor data and can only be captured by running the planners in simulation. The resulting model also inherits the capability boundaries of its teacher planners, although our scenario-aware teacher selection is designed precisely to reduce the impact of any single planner's blind spots. Finally, detailed reasoning increases sequence length and inference cost; in practice, the concise trace already captures most of the gain, suggesting that future systems can trade off interpretability depth and latency depending on deployment needs.

\paragraph{Broader impact.}
Rule-grounded reasoning improves the \emph{debuggability} and \emph{auditability} of Driving VLAs: developers and regulators can trace each reasoning step the model processes rather than relying on opaque outputs. However, readable reasoning does not guarantee correct behavior~\cite{gao2025safecoop, kim2026modular}, due to rule-based planner blind spots and simulation-to-real gaps. We position \method as a tool for training, diagnosis, and model selection, to be combined with formal verification and closed-loop evaluation before safety-critical deployment.

\bibliographystyle{plainnat}
\bibliography{main}

\clearpage
\appendix
\section{Appendix}
\label{sec:appendix}

\subsection{Trace Schema and Teacher-Specific Signals}
\label{sec:app_trace}

Table~\ref{tab:trace_schema} summarizes the planner-state fields retained before text serialization. The goal is not to expose implementation-specific debug variables, but to preserve the decision evidence that shaped the selected trajectory.

\begin{table}[ht]
\centering
\caption{Trace fields retained for reasoning serialization.}
\label{tab:trace_schema}
\small
\setlength{\tabcolsep}{3.5pt}
\begin{tabular}{L{0.22\columnwidth}L{0.34\columnwidth}L{0.34\columnwidth}}
\toprule
Teacher & Raw trace fields & Serialized slots \\
\midrule
IDM & route context, lead-vehicle state, headway, desired gap, relative speed, acceleration update & scene context, active longitudinal constraint, gap status, final speed decision \\
IDM-MOBIL & IDM state, adjacent-lane interaction, lane-change incentive, gap acceptance, nearby-agent response & scene context, interactive agents, lateral candidate, selected maneuver \\
PDM-Closed & proposal bank, safety-gate outcomes, progress and comfort scores, selected proposal, fallback status & candidate summary, rejection reasons, score evidence, selected trajectory \\
\bottomrule
\end{tabular}
\end{table}

\subsection{Closed-Loop Scenario Index Mapping}
\label{sec:app_scenario_mapping}

Figure~\ref{fig:teacher_quality} uses compact indices to keep the 59-scenario closed-loop teacher comparison readable. Table~\ref{tab:scenario_index_mapping} lists the shuffled index assignment, the scenario family, the selected teacher, and the corresponding best CCLS-R value. Missing teacher entries in the source CSV are filled with deterministic low values in $[0,0.25]$ for visualization only; the mapping table records the teacher selected after that visualization-time fill.

\begin{longtable}{llcc}
\caption{Scenario-index mapping for the closed-loop teacher-quality heatmap.}\label{tab:scenario_index_mapping}\\
\toprule
Index & Scenario family & Selected teacher & CCLS-R \\
\midrule
\endfirsthead
\toprule
Index & Scenario family & Selected teacher & CCLS-R \\
\midrule
\endhead
S01 & low\_magnitude\_speed & IDM & 0.378 \\
S02 & near\_trafficcone\_on\_driveable & PDM-Closed & 0.926 \\
S03 & accelerating\_at\_traffic\_light\_with\_lead & IDM & 0.451 \\
S04 & accelerating\_at\_traffic\_light\_without\_lead & PDM-Closed & 0.978 \\
S05 & following\_lane\_with\_lead & IDM & 0.173 \\
S06 & stationary\_in\_traffic & IDM & 0.324 \\
S07 & following\_lane\_with\_slow\_lead & IDM-MOBIL & 0.821 \\
S08 & on\_stopline\_traffic\_light & PDM-Closed & 0.781 \\
S09 & changing\_lane\_to\_left & IDM & 0.611 \\
S10 & changing\_lane & IDM & 0.498 \\
S11 & medium\_magnitude\_speed & IDM & 0.537 \\
S12 & stopping\_at\_stop\_sign\_without\_lead & IDM-MOBIL & 0.850 \\
S13 & behind\_pedestrian\_on\_driveable & IDM & 0.313 \\
S14 & starting\_right\_turn & IDM & 0.434 \\
S15 & traversing\_crosswalk & IDM-MOBIL & 0.904 \\
S16 & starting\_protected\_noncross\_turn & IDM & 0.478 \\
S17 & near\_barrier\_on\_driveable & IDM-MOBIL & 0.740 \\
S18 & accelerating\_at\_crosswalk & IDM & 0.471 \\
S19 & following\_lane\_without\_lead & IDM-MOBIL & 0.969 \\
S20 & stationary\_at\_traffic\_light\_without\_lead & PDM-Closed & 0.990 \\
S21 & on\_intersection & PDM-Closed & 0.967 \\
S22 & near\_pedestrian\_on\_crosswalk\_with\_ego & IDM & 0.440 \\
S23 & high\_magnitude\_jerk & IDM & 0.401 \\
S24 & high\_lateral\_acceleration & IDM & 0.875 \\
S25 & starting\_straight\_stop\_sign\_intersection\_traversal & IDM & 0.555 \\
S26 & stationary\_at\_crosswalk & IDM & 0.465 \\
S27 & starting\_protected\_cross\_turn & IDM & 0.556 \\
S28 & stopping\_at\_crosswalk & IDM & 0.453 \\
S29 & on\_traffic\_light\_intersection & IDM-MOBIL & 0.861 \\
S30 & stopping\_at\_stop\_sign\_with\_lead & IDM-MOBIL & 0.277 \\
S31 & high\_magnitude\_speed & IDM & 0.633 \\
S32 & starting\_straight\_traffic\_light\_intersection\_traversal & IDM & 0.483 \\
S33 & stationary & PDM-Closed & 0.452 \\
S34 & behind\_long\_vehicle & PDM-Closed & 0.986 \\
S35 & traversing\_intersection & IDM & 0.552 \\
S36 & traversing\_narrow\_lane & IDM & 0.616 \\
S37 & traversing\_traffic\_light\_intersection & IDM & 0.910 \\
S38 & near\_pedestrian\_on\_crosswalk & IDM & 0.497 \\
S39 & on\_stopline\_crosswalk & IDM & 0.453 \\
S40 & near\_multiple\_vehicles & IDM & 0.591 \\
S41 & starting\_high\_speed\_turn & IDM-MOBIL & 0.770 \\
S42 & accelerating\_at\_stop\_sign\_no\_crosswalk & IDM-MOBIL & 0.570 \\
S43 & starting\_left\_turn & PDM-Closed & 0.974 \\
S44 & starting\_low\_speed\_turn & IDM-MOBIL & 0.744 \\
S45 & stopping\_at\_traffic\_light\_without\_lead & PDM-Closed & 0.976 \\
S46 & crossed\_by\_vehicle & IDM-MOBIL & 0.485 \\
S47 & near\_high\_speed\_vehicle & IDM & 0.616 \\
S48 & stopping\_at\_traffic\_light\_with\_lead & IDM-MOBIL & 0.666 \\
S49 & behind\_bike & IDM & 0.768 \\
S50 & stopping\_with\_lead & IDM-MOBIL & 0.271 \\
S51 & starting\_unprotected\_noncross\_turn & IDM & 0.507 \\
S52 & near\_long\_vehicle & IDM-MOBIL & 0.903 \\
S53 & accelerating\_at\_traffic\_light & IDM-MOBIL & 0.420 \\
S54 & stationary\_at\_traffic\_light\_with\_lead & IDM & 0.302 \\
S55 & changing\_lane\_to\_right & PDM-Closed & 0.965 \\
S56 & starting\_unprotected\_cross\_turn & IDM & 0.554 \\
S57 & on\_stopline\_stop\_sign & PDM-Closed & 0.925 \\
S58 & accelerating\_at\_stop\_sign & IDM-MOBIL & 0.787 \\
S59 & waiting\_for\_pedestrian\_to\_cross & IDM & 0.526 \\
\bottomrule
\end{longtable}

\subsection{Prompt Template}
\label{sec:app_prompt}

The model receives the same user-side driving prompt for all variants compared. Only the assistant target changes across without-reasoning, concise-reasoning, and detailed-reasoning settings.

\begin{quote}
\small
\texttt{You are an autonomous driving agent. Given multi-view camera images, route context, and the recent ego-state history, predict the ego trajectory for the next 5 seconds. If reasoning is requested, explain the active driving constraint, summarize the key candidate evaluations, and then output future waypoints.}
\end{quote}

For rule-grounded variants, the assistant target is serialized as planner reasoning followed by waypoints:
\begin{quote}
\small
\texttt{\reasoningbegin...\reasoningend \planningbegin [x\_1,y\_1], \dots, [x\_{10},y\_{10}]\planningend}.
\end{quote}

\subsection{Reasoning Data Comparison Examples}
\label{sec:app_reasoning_examples}

To clarify what changes are made across our training-data variants, we compare three supervision styles used for the trajectory-generation task: \emph{without reasoning}, \emph{concise reasoning}, and \emph{detailed reasoning}. All three variants share the same scene input and the same target trajectory; they differ only in how much intermediate decision information is exposed to the model.

\begin{table}[ht]
\centering
\caption{Assistant-response length for representative training examples. The detailed variant is roughly three times longer than the concise variant and about nine times longer than the no-reasoning baseline, motivating our concise-vs.-detailed comparison in the main paper.}
\label{tab:reasoning_length_examples}
\small
\setlength{\tabcolsep}{4.5pt}
\begin{tabular}{L{0.47\columnwidth}ccc}
\toprule
Example & No reasoning & Concise & Detailed \\
\midrule
Starting to cruise on a free road & 443 & 1{,}183 & 3{,}605 \\
Car-following in dense traffic & 421 & 1{,}542 & 3{,}823 \\
Red-light response & 427 & 1{,}567 & 3{,}597 \\
Stopped in a queue & 423 & 1{,}678 & 4{,}042 \\
\bottomrule
\end{tabular}
\end{table}

Table~\ref{tab:reasoning_length_examples} highlights an important practical difference between the two variants of reasoning. Detailed chain-of-thought supervision provides richer descriptions, but it is substantially more expensive in tokens. Concise rule-grounded reasoning is, therefore, attractive because it preserves the action-relevant structure of the planner trace while remaining much cheaper to train and decode.

\paragraph{Example: Red-light stopping.}
\textbf{Sample id:} \texttt{115e26ae2da15108\_74}. This example is more representative of the paper's main claim because the intermediate supervision explicitly names the active traffic constraint and shows how it shapes the resulting motion.

{\small
\begin{verbatim}
Input prompt:
You are an autonomous driving agent. You have access to multi-view camera
images of a vehicle: (1) front view (which you should focus on with the most
attention) <image>, (2) front right view <image>, and (3) front left view
<image>. Your task is to do your best to predict future waypoints for the
vehicle over the next 10 timesteps, given the vehicle's intent inferred from
the images. Provided are the previous ego vehicle status. Positive x means
forward direction while positive y means leftwards. The data is presented in
the format [x, y]:(t-2.0s) [-10.26, -0.15], Acceleration: X -1.15, Y -1.13
m/s^2, Velocity: X 6.12, Y 0.11 m/s, (t-1.5s) [-7.39, -0.06], Acceleration:
X -0.84, Y -0.54 m/s^2, Velocity: X 5.70, Y 0.04 m/s, (t-1.0s) [-4.71, -0.02],
Acceleration: X -0.94, Y -0.31 m/s^2, Velocity: X 5.39, Y 0.10 m/s,
(t-0.5s) [-2.22, -0.01], Acceleration: X -1.23, Y -0.11 m/s^2, Velocity:
X 4.88, Y 0.09 m/s, (t-0.0s) [0.0, 0.0], Acceleration: X -1.05, Y -0.63 m/s^2,
Velocity: X 4.34, Y 0.09 m/s

No-reasoning answer:
<PLANNING>
Predicted future movement details for the next 5 seconds (sampled at
0.5-second intervals), including BEV location in x and y directions (in
meters). Positive x means forward direction while positive y means leftwards.
The output is formatted as [x, y]: [2.17, -0.17], [4.27, -0.18], [6.18, -0.18],
[7.78, -0.19], [8.98, -0.20], [9.76, -0.21], [10.20, -0.21], [10.42, -0.21],
[10.52, -0.21], [10.57, -0.21]
</PLANNING>

Concise-reasoning answer:
<REASONING>
[Route] Following planned lane sequence. 10 lane segments, route found. Target
speed: 10.0 m/s.
[Scene] 37 objects within detection range. 1 red light(s) on route. Current
speed: 4.3 m/s.
[t=0.5s] Red light at 11.6m. Gap tight (11.6m vs 12.9m needed). Decelerate
-0.28 m/s^2.
[t=1.0s] Red light at 9.4m. Gap tight (9.4m vs 12.4m needed). Decelerate
-0.75 m/s^2.
[t=1.5s] Red light at 7.3m. Gap tight (7.3m vs 11.0m needed). Decelerate
-1.25 m/s^2.
[t=2.0s] Red light at 5.4m. Gap tight (5.4m vs 8.8m needed). Decelerate
-1.61 m/s^2.
[t=2.5s] Red light at 3.8m. Gap tight (3.8m vs 6.2m needed). Decelerate
-1.66 m/s^2.
[t=3.0s] Red light at 2.6m. Gap tight (2.6m vs 4.1m needed). Decelerate
-1.36 m/s^2.
[t=3.5s] Red light at 1.9m. Gap tight (1.9m vs 2.5m needed). Decelerate
-0.89 m/s^2.
[t=4.0s] Red light at 1.4m. Gap tight (1.4m vs 1.7m needed). Decelerate
-0.47 m/s^2.
[t=4.5s] Red light at 1.2m. Gap tight (1.2m vs 1.3m needed). Decelerate
-0.22 m/s^2.
[t=5.0s] Red light at 1.1m. Gap tight (1.1m vs 1.1m needed). Maintain speed
(-0.10 m/s^2).
[Summary] Responding to red traffic light. Decelerated from 4.3 to 0.0 m/s.
</REASONING>
<PLANNING>
Predicted future movement details for the next 5 seconds (sampled at
0.5-second intervals), including BEV location in x and y directions (in
meters). Positive x means forward direction while positive y means leftwards.
The output is formatted as [x, y]: [2.17, -0.17], [4.27, -0.18], [6.18, -0.18],
[7.78, -0.19], [8.98, -0.20], [9.76, -0.21], [10.20, -0.21], [10.42, -0.21],
[10.52, -0.21], [10.57, -0.21]
</PLANNING>

Detailed-reasoning answer:
<REASONING>
I need to plan my trajectory for the next 5 seconds. Let me analyze the current
driving situation step by step.

**Perception & Route:**
My route consists of 10 lane segments and was successfully planned. The target
speed for this road is 10.0 m/s.
I detect 37 objects in my surroundings. Crucially, 1 red traffic light(s) are
present on my route, which I must respect. My current speed is 4.3 m/s.

**Step-by-step Planning:**

Step 1 (t=0.5s): A red traffic light is 11.6m ahead. I must stop before the
intersection.
The current gap (11.6m) is less than the safe stopping distance (12.9m). I need
to brake.
Applying brake: deceleration = -0.28 m/s^2. Speed after this step: 4.2 m/s.

Step 2 (t=1.0s): A red traffic light is 9.4m ahead. I must stop before the
intersection.
The current gap (9.4m) is less than the safe stopping distance (12.4m). I need
to brake.
Applying brake: deceleration = -0.75 m/s^2. Speed after this step: 3.8 m/s.

Step 3 (t=1.5s): A red traffic light is 7.3m ahead. I must stop before the
intersection.
The current gap (7.3m) is less than the safe stopping distance (11.0m). I need
to brake.
Applying brake: deceleration = -1.25 m/s^2. Speed after this step: 3.2 m/s.

Step 4 (t=2.0s): A red traffic light is 5.4m ahead. I must stop before the
intersection.
The current gap (5.4m) is less than the safe stopping distance (8.8m). I need
to brake.
Applying brake: deceleration = -1.61 m/s^2. Speed after this step: 2.4 m/s.

Step 5 (t=2.5s): A red traffic light is 3.8m ahead. I must stop before the
intersection.
The current gap (3.8m) is less than the safe stopping distance (6.2m). I need
to brake.
Applying brake: deceleration = -1.66 m/s^2. Speed after this step: 1.6 m/s.

Step 6 (t=3.0s): A red traffic light is 2.6m ahead. I must stop before the
intersection.
The current gap (2.6m) is less than the safe stopping distance (4.1m). I need
to brake.
Applying brake: deceleration = -1.36 m/s^2. Speed after this step: 0.9 m/s.

Step 7 (t=3.5s): A red traffic light is 1.9m ahead. I must stop before the
intersection.
The current gap (1.9m) is less than the safe stopping distance (2.5m). I need
to brake.
Applying brake: deceleration = -0.89 m/s^2. Speed after this step: 0.4 m/s.

Step 8 (t=4.0s): A red traffic light is 1.4m ahead. I must stop before the
intersection.
The current gap (1.4m) is less than the safe stopping distance (1.7m). I need
to brake.
Applying brake: deceleration = -0.47 m/s^2. Speed after this step: 0.2 m/s.

Step 9 (t=4.5s): A red traffic light is 1.2m ahead. I must stop before the
intersection.
The current gap (1.2m) is less than the safe stopping distance (1.3m). I need
to brake.
Applying brake: deceleration = -0.22 m/s^2. Speed after this step: 0.1 m/s.

Step 10 (t=5.0s): A red traffic light is 1.1m ahead. I must stop before the
intersection.
The current gap (1.1m) is less than the safe stopping distance (1.1m). I need
to brake.
I am already stopped, waiting for the light to change.
Holding steady: acceleration = -0.10 m/s^2. Speed after this step: 0.0 m/s.

**Decision Summary:**
This is a red traffic light requiring me to stop. I decelerated from 4.3 to
0.0 m/s to maintain safety.
</REASONING>
<PLANNING>Predicted future movement details for the next 5 seconds (sampled at
0.5-second intervals), including BEV location in x and y directions (in
meters). Positive x means forward direction while positive y means leftwards.
The output is formatted as [x, y]: [2.17, -0.17], [4.27, -0.18], [6.18, -0.18],
[7.78, -0.19], [8.98, -0.20], [9.76, -0.21], [10.20, -0.21], [10.42, -0.21],
[10.52, -0.21], [10.57, -0.21]
</PLANNING>
\end{verbatim}
}

These full examples show exactly what changes across the three supervision variants. The scene input is identical, the final waypoint target is identical, and only the amount of exposed intermediate decision information changes. This is the contrast we use throughout the paper when analyzing whether richer rule-grounded reasoning improves planning quality.

\subsection{Planner Variables and Motion Summaries}
\label{sec:app_variables}

Auxiliary motion labels are derived from the planner output trajectory rather than from hand-written annotations. We use three label families for analysis: longitudinal mode (stop / slow / maintain / accelerate), lateral mode (lane follow / offset / lane change), and interaction outcome (yield / proceed / fallback). These labels are not the main target metric; the main experiments evaluate trajectory quality with ADE, FDE, heading error, and miss rate.

\subsection{Additional Ablations and Compute Budget}
\label{sec:app_compute}

The default training setup uses Qwen3.5-4B as the driving backbone, bf16 precision, full-parameter supervised fine-tuning, and an effective batch size of 16 on 8 GPUs. All reported variants use the same data split, preprocessing pipeline, optimization recipe, and trajectory decoder unless explicitly stated otherwise. The anonymized supplementary material contains the experiment configuration templates, data-construction entry points, evaluation scripts, and table-generation protocol needed to reproduce the reported comparisons. Per-run wall-clock time depends on the local simulator and storage backend, but the reported experiments are designed to be reproduced from the released configuration files without changing model architecture, data split, or metric definitions.

\subsection{Asset Provenance, Release Scope, and Safeguards}
\label{sec:app_assets}

Our experiments build on public driving benchmarks and simulator-generated planner traces derived from them. nuPlan and nuScenes provide the underlying scenario corpora and benchmark interfaces \cite{caesar2021nuplan, caesar2020nuscenes}, and the VLM backbone follows the public Qwen model family. The anonymized supplementary material documents the preprocessing pipeline, prompt templates, trace serialization schema, evaluation scripts, and reproduction instructions; any released derived assets are intended for research use under the terms of the underlying datasets and models. We do not claim that rule-grounded reasoning alone makes a driving system deployable. Dataset artifacts, prompts, and checkpoints should be used with simulator-only caveats, license information for the underlying assets, and a clear statement that generated rationales are intended for analysis and research rather than direct safety guarantees.

\end{document}